\pdfoutput=1
\documentclass[10pt, a4paper]{article}

\usepackage{hyperref}
\usepackage{breakurl}

\usepackage{lrec-coling2024} % this is the new style

\usepackage{comment}
\usepackage{booktabs}
\usepackage{linguex}
\usepackage{tikz}
\usepackage{tikz-qtree}
\usepackage{tikz-dependency}
\usetikzlibrary{math,calc,positioning,shapes.symbols,fit,quotes,arrows.meta,patterns,graphs,shapes.callouts}
\usepackage{pgfplots}
\pgfplotsset{scaled y ticks=false}
\tikzset{>=latex}

\usepackage{amssymb}
\usepackage{amsthm}
\usepackage[framemethod=TikZ]{mdframed}
\definecolor{definitioncolor}{rgb}{0.94, 0.97, 1.0}  % lightblue
\mdfsetup{
    backgroundcolor=definitioncolor,
    linewidth=1pt,
    roundcorner=5pt
}
\theoremstyle{definition}
\newmdtheoremenv{definition}{Definition}

\newcommand{\ensuretext}[1]{#1}
\newcommand{\nertcomment}[4]{\ensuretext{\textcolor{#3}{[\ensuretext{\textcolor{#3}{\ensuremath{^{\textsc{#1}}_{\textsc{#2}}}}} #4]}}}
\newcommand{\sh}[1]{\nertcomment{S}{H}{magenta}{#1}}
\newcommand{\sk}[1]{\nertcomment{S}{K}{purple}{#1}}
\newcommand{\cc}[1]{\nertcomment{C}{C}{orange}{#1}}

\newcommand{\affil}[1]{$^\textnormal{{#1}}$}

\usepackage{amsmath,amsfonts,bm}

\def\eqref#1{equation~\ref{#1}}
\def\1{\bm{1}}

\def\vzero{{\bm{0}}}

\def\va{{\bm{a}}}

\def\vx{{\bm{x}}}

\def\eva{{a}}

\def\evx{{x}}

\DeclareMathAlphabet{\mathsfit}{\encodingdefault}{\sfdefault}{m}{sl}
\SetMathAlphabet{\mathsfit}{bold}{\encodingdefault}{\sfdefault}{bx}{n}

\newcommand{\R}{\mathbb{R}}

\title{
Sparse Logistic Regression with High-order Features\\
for Automatic Grammar Rule Extraction from Treebanks
}

\name{Santiago Herrera\affil{1}, Caio Corro\affil{2}, Sylvain Kahane\affil{1,3}} 

\address{\affil{1}Université Paris Nanterre, CNRS, Modyco~~~\affil{2}Sorbonne Université, CNRS, ISIR \\
\affil{3}Institut Universitaire de France\\
         sherrera@parisnanterre.fr, caio.corro@isir.upmc.fr, sylvain@kahane.fr\\
         }

\abstract{
Descriptive grammars are highly valuable, but writing them is time-consuming and difficult. Furthermore, while linguists typically use corpora to create them, grammar descriptions often lack quantitative data. As for formal grammars, they can be challenging to interpret. In this paper, we propose a new method to extract and explore significant fine-grained grammar patterns and potential syntactic grammar rules from treebanks, in order to create an easy-to-understand corpus-based grammar. More specifically, we extract descriptions and rules across different languages for two linguistic phenomena, agreement and word order, using a large search space and paying special attention to the ranking order of the extracted rules. For that, we use a linear classifier to extract the most salient features that predict the linguistic phenomena under study. We associate statistical information to each rule, and we compare the ranking of the model's results to those of other quantitative and statistical measures. Our method captures both well-known and less well-known significant grammar rules in Spanish, French, and Wolof.\\ \newline 
}

\begin{document}

\maketitleabstract

\section{Introduction}\label{sec:intro}

\begin{figure*}[ht]
\centering
\begin{tikzpicture}[
    every node/.style={
        rectangle,
        inner xsep=0cm,
        inner ysep=0.1cm,
        text height=1.5ex,
        text depth=.25ex,
        node distance=0.5cm
    },
    word/.append style={
        font=\tt\normalsize,
    },
    deplabel/.append style={
        font=\sc\footnotesize,
        above
    },
    upos/.append style={
        node distance=0cm,
        yshift=0.1cm,
        font=\sc\footnotesize
    }
]

    \node [word] (it) {It};
    \node [upos, below=of it] {pron};
    
    \node [word] (follows) [right=of it] {follows};
    \node [upos, below=of follows] {verb};
    
    \node [word] (that) [right=of follows] {that};
    \node [upos, below=of that] {sconj};
    
    \node [word] (thea) [right=of that] {the};
    \node [upos, below=of thea] {det};
    
    \node [word] (role) [right=of thea] {role};
    \node [upos, below=of role] {noun};
    
    \node [word] (of) [right=of role] {of};
    \node [upos, below=of of] {adp};
    
    \node [word] (theb) [right=of of] {the};
    \node [upos, below=of theb] {det};
    
    \node [word] (state) [right=of theb] {state};
    \node [upos, below=of state] {noun};
    
    \node [word] (is) [right=of state] {is};
    \node [upos, below=of is] {aux};

    \node [word] (essential) [right=of is] {essential};
    \node [upos, below=of essential] {adj};

    \draw [->]
        ($(follows.north)+(-3pt,0)$)
        |-
        ($(follows.north)!0.5!(it.north)+(0,0.3)$)
        node[deplabel,xshift=-0.2cm] {comp@expl}
        -|
        (it.north)
    ;
    \draw [->]
        ($(follows.north)+(+0.5pt,0)$)
        |-
        ($(follows.north)!0.5!(that.north)+(0,0.3)$)
        node[deplabel,xshift=0cm] {subj}
        -|
        (that.north)
    ;
    \draw [->]
        ($(that.north)+(+3pt,0)$)
        |-
        ($(that.north)!0.5!(is.north)+(0,1.8)$)
        node[deplabel] {comp:obj}
        -|
        (is.north)
    ;
    \draw [->]
        ($(is.north)+(-4pt,0)$)
        |-
        ($(is.north)!0.5!(role.north)+(0,1.3)$)
        node[deplabel] {subj}
        -|
        (role.north)
    ;
    \draw [->]
        ($(role.north)+(-3pt,0)$)
        |-
        ($(role.north)!0.5!(thea.north)+(0,0.3)$)
        node[deplabel,xshift=-0.1cm] {det}
        -|
        (thea.north)
    ;
    \draw [->]
        ($(role.north)+(+3pt,0)$)
        |-
        ($(role.north)!0.5!(of.north)+(0,0.3)$)
        node[deplabel,xshift=0.1cm] {udep}
        -|
        (of.north)
    ;
    
    \draw [->]
        ($(of.north)+(3pt,0)$)
        |-
        ($(of.north)!0.5!(state.north)+(0,0.8)$)
        node[deplabel] {comp:obj}
        -|
        (state.north)
    ;
    \draw [->]
        ($(state.north)+(-3pt,0)$)
        |-
        ($(state.north)!0.5!(theb.north)+(0,0.3)$)
        node[deplabel,xshift=-0.1cm] {det}
        -|
        (theb.north)
    ;
    
    \draw [->]
        ($(is.north)+(4pt,0)$)
        |-
        ($(is.north)!0.5!(essential.north)+(0,0.3)$)
        node[deplabel,,xshift=0.1cm] {comp:pred}
        -|
        (essential.north)
    ;

\end{tikzpicture}
    \caption{Example from the SUD version of the English GUM treebank. Two clauses with subjects in different positions. In the main clause, the subject follows the verb and the first position is filled by an expletive. In the subordinate clause, the subject occupies the dominant pre-verbal position.}
    \label{fig:subj_parataxis}

\end{figure*}
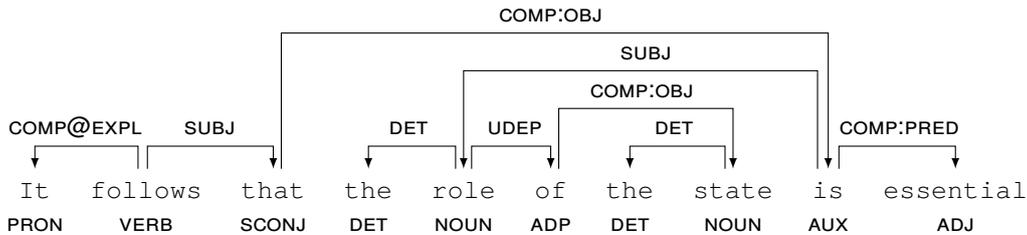

Speakers are obliged to select, pronounce, use, mark, and combine linguistic units in precise ways in order to communicate in a language.
A grammar is this set of constraints that a language imposes on its speakers, \emph{i.e.}\ a set of constraints satisfied by all well-formed utterances.
In this setting, a grammar rule describes a specific linguistic pattern enforced in a given context and in a given language, and can be of various interest (\emph{e.g.}\ education, theoretical linguistics, linguistic typology).
In practice, these rules are of probabilistic nature (in the frequentist interpretation).
For example, in English, a simple rule would be that the ``\emph{subject of a verb is before its governor}''.
However, this rule is not deterministic, a more accurate description would be: ``\emph{given a syntactic dependency of type subject, the dependent is before its governor in approximately 93\% of cases}''.
In practice, it is of interest to understand when this majority rule is not followed.\footnote{The value also depends on the annotation scheme and on the corpus considered. Here the results are given for SUD\_English-GUM@2.13, converted from UD in SUD \cite{gerdes2018sud}, where the subject depends on the auxiliary in complex verbal forms.}
To continue with the same example: ``\emph{given a syntactic dependency of type subject with an expletive complement in a pre-verbal position, the subject is after its governor in 97\% of cases}'', see Figure~\ref{fig:subj_parataxis}.
%As such, rules are not only probabilistic but ranked in relation to each other.
As such, rules are not only probabilistic but also more or less fine-grained and potentially overlapping.
%\cc{J'avais utilisé hierarchie plutôt que ranked, je ne suis pas trop sur de vouloir utiliser ranked ici...}

With the development of large treebank collections like Universal Dependencies,
there has been a recent interest in automatically extracting grammars from syntactically annotated sentences.
Importantly, previous work by \citet{chaudhary-etal-2020-automatic,chaudhary2022autolex} proposes to rely on machine learning techniques to this end.
However, their approach suffers from some limitations, which we overcome in our work:
(1) their syntactic rules do not capture fine-grained phenomena due to an under-specified search space and the use of decision trees (\emph{e.g.}\ their model capture the fact that the subject inversion can be due to the presence of 'there', but not by the presence of an expletive in general, missing the example of Figure~\ref{fig:subj_parataxis});
(2) the use of decision trees \cite{quinlan1986induction} requires tuning of several hyperparameters (entropy choice, depth, splitting constraint, etc.), which is difficult to do on held-out data due to the frequentist nature of predictions;
(3) a decision tree does not produce a ranked list of rules by saliency.

In this paper, we propose a different method to extract and explore more fine-grained grammar rules from syntactic treebanks
in order to automatically create an easy-to-understand corpus-based grammar.
More specifically, we extract descriptions across languages for two linguistic phenomena, agreement and word order, paying special attention to the ranking order of the extracted rules.
To that end, we use a linear classifier trained with a sparsity inducing regularizer \cite{bach2012optimization}.
The strength of the regularization, denoted $\lambda \in \R_+$, directly impacts the number of extracted rules for the linguistic phenomena under study.
In this setting, the regularization path is the sequence of solutions to the training problem when $\lambda$ varies \cite{markowitz1952portfolio,osborne2000lasso,efron2004least}.
The earlier a rule appears in the path when decreasing $\lambda$, the salient it is for the linguistic phenomena under study.
Another important benefit of our approach compared to decision trees is that it is (almost) hyperparameter free, which is a crucial feature as hyperparameter selection is non-trivial (and even hopeless), see Section~\ref{sec:reg_path}.

Our contributions can be summarized as follows:
\begin{itemize}
    \item we propose a novel formalization of grammatical rules with the aim of automatic extraction from treebanks;
    \item compared to previous work, we broaden the expressiveness of patterns to allow for more precise rules;
    \item we investigate the use of sparse logistic regression to extract and rank rules;
    \item we conduct qualitative evaluation on three languages (Spanish, French and Wolof).
\end{itemize}
All in all, we contribute to the development of descriptive grammars that, while time-consuming to write, are fundamental resources for theoretical linguistic, language-specific or cross-linguistic studies, and an essential resource for language documentation. Our goal is to contribute to the brisdging of the gap between computational linguistics (CL) and theoretical linguistics \citep{baroni_on_the_gap2021}, as well as between CL and field linguistics.

\section{Related Works}

\textbf{Formal grammars (FG).} There exists a large amount of works on extracting FGs from treebanks,
including CFG \citep{charniak1996treebankgrammar},
TAG \citep{chen_bangalore_vijay-shanker_2006,bhatt-etal-2012-tag}, LFG \citep{frank2003lfg,burke2004treebank,rehbein2009treebank}
and HPSG \citep{cahill2005treebank,zhang2012hpsg} grammars.
FGs are rich and complete descriptions of structures observed in treebanks, usually intended for parsing.
However, they are also redundant and difficult to interpret as they are not descriptive grammars \citep{zaefferer1992framework} and because of their very large size.\footnote{\citeauthor{charniak1996treebankgrammar}'s \citeyearpar{charniak1996treebankgrammar} English grammar contains $15\,953$ rules.}
Moreover, these FG have limitations when it comes to incorporating quantitative information and highlighting salient features of corpora as they are based on strong symbolic formalisms.
%They could easily encode how likely a grammatical structure is, but not how salient it is relative to other structures.
%\cc{J'ai commenté le paragraphe suivant (voir latex) sur marsagram car trop verbeux. J'ai ajouté la citation à marsagram ci-dessus, mais du coup il nous manque alves-etal-2023-analysis.}

\begin{comment}
In \citet{blache-etal-2016-marsagram}, in addition to extracting a CFG grammar from treebanks, they compute some meta-linguistic properties over each extracted LHS and RHS pair \sk{?}, being the nodes as deprel-POS features. These meta-properties include information about the exclusion or requirement of two nodes, the uniqueness of nodes in a syntactic group, or the linear order between nodes. These grammars have proven to be interesting for computing typological familiarity between languages \citep{blache-etal-2016-marsagram, alves-etal-2023-analysis}. However, since it is still a CFG, this grammar is redundant from a descriptive point of view, and limited in that it extracts only formal rules of depth 1 (between a node and its immediate constituents). Indeed, it is more computationally expensive to extend the search space to include other nodes and the morphological features encoded in a treebank. However, this approach falls short in capturing fine-grained grammar rules. 
\end{comment}

\textbf{Typologicial features induction.}
Although linguists still rely on databases like WALS \citep{wals} and GramBank \citep{grambank_release} for typological features, it is becoming increasingly important to infer these features using corpora, see \citet{ponti-etal-2019-modeling} for motivations.
Importantly, there has been a recent interest on gradient and quantitative feature descriptions \cite{levshina2022, baylor-etal-2024-multilingual} instead of the standard categorical setting.
%especially in a gradient and quantitative way \cite{levshina2022, baylor-etal-2024-multilingual} rather than in categorical way.
%The use of corpora to extract not only grammatical but also typological features in a cross-linguistic manner has become increasingly common in recent years.
For examples, previous work have considered formal grammars parametrized by categorical typological features, parameters which are inferred from various types of corpora, including interlinear glosses \citep{bender-etal-2014-learning,howell-bender-2022-building}. Similarly, \citet{choi-etal-2021-corpus,choi-etal-2021-investigating} used manually crafted rules to extract typological statistics from treebanks, whereas \citet{alves-etal-2023-analysis} also relied on information from an automatically extracted CFG. Contrary to these works, our aim is to provide more fine-grained and quantitative grammar rules while requiring less manual pattern crafting.

\textbf{Grammar rule extraction.}
Closer to our work, \citet{chaudhary-etal-2020-automatic, chaudhary2022autolex} demonstrate the ability to extract human-readable grammar rules using machine learning techniques.
The authors focus on a few specific linguistic phenomena, including agreement and word order, that are cast as classification tasks.
For example, they investigate the subset of features (encoded in governor-dependent pairs) used by a decision tree to predict if there is a number agreement between two dependent words.
%While this method provides concise and readable rules, they are not quantitatively rankable and they are note expressive enough to encode many important phenomena.
Finally, \citet{blache-etal-2016-marsagram} investigate the use of an automatically extracted CFG with the same goal.
Their approach also suffers from the expressiveness issue due to the locality of CFG's rewriting rules.
%Additionally, the evaluation does not inform us about the quality of this grammar compared to other extracted grammars or how much simpler or complex it is.

%To evaluate the optimal set of grammar patterns internally, \citet{dunn-2019-frequency} used the Minimum Description Length (MDL) principle to simultaneously assess how well a set of extracted constructions covers the dataset and how simple this set of constructions is. The MDL principle has recently proven to be useful in linguistic contexts. \citet{kroos2022difference} applies sequential data mining techniques in combination with the MDL principle to extract linguistic building blocks from corpora's POS sequences. A new graph mining algorithm based on MDL \citep{bariatti2020graphmdl} was tested on UD treebanks.

\section{Grammatical Formalism}
\label{sec:formalism}

\subsection{Syntactic Dependencies}
\begin{table}[t!]
\centering
\begin{tabular}{lrr}
\toprule
\textbf{Treebank} & \textbf{Sentences} & \textbf{Tokens}\\
\midrule
English-GUM & 10\,k & 187\,k \\
French-GSD & 16\,k & 400\,k \\
%French-Rhapsodie & 3\,k & 44\,k \\
Spanish-AnCora & 17\,k & 559\,k\\
Wolof-WTB & 2\,k & 44\,k\\
\bottomrule
\end{tabular}
\caption{Treebank sizes.}
\label{table:1}
\end{table}

Although our method for extracting rules does not depend on a particular formalism, we briefly present here the data to which the method has been applied, that is, the Universal Dependencies project \citep[UD,][]{demarneffe2021universal}. It is an open project on cross-linguistic syntactic annotation, which has 259 treebanks for 148 different languages in its last available release (UD 2.13).
It is based on a consistent framework for multilingual annotation of parts of speech, lemmas, morphological features and syntactic dependencies, as well as guidelines for lemmatization, tokenization, and other tasks.
%UD allows the annotation of other type of linguistic information, as long as it respects the constraints imposed by the CONLL-U format used to encode the treebanks.

UD defines content words as syntactic heads in order to make annotations more seamless between different languages.
While this favors comparability between treebanks, some morphosyntactic properties remain ``less accessible'' and syntactic differences less evident.
Given that extracted rules depends on the annotation scheme,
we instead use the Surface-Syntactic UD (SUD) framework \citep{gerdes2018sud, gerdes-2021}, which defines syntactic heads by distributional criteria.
This is more suitable to extract word order and agreement rules, amongst others.
For example, in SUD, contrary to UD, auxiliaries are analyzed as the head of the clause and the subject depends on it.
In English (and more generally in Indo-European languages), the subject is positioned towards the auxiliary, with which it agrees, rather than the verb.
Moreover, this allows as to compare our work to previous work \citep{chaudhary-etal-2020-automatic, chaudhary2022autolex} that also use SUD.

In order to demonstrate how our method works, we have chosen to analyze the results on languages in which we are experts. This allows us to conduct precise qualitative analysis of results, which is in our opinion more important that large scale evaluation via crowdsourcing native speakers.\footnote{There are evidences that crowd workers automatize their tasks using large language models \cite{veselovsky2023artificial}, which make these evaluations less reliable.}
%We work on languages we have a native or proficiency level to evaluate the extracted rules.
In section~\ref{sec:results}, we present rules extracted from SUD\_Spanish-AnCora \citep{taule-etal-2008-ancora}, SUD\_French-GSD \citep{guillaume_2019}, and SUD\_Wolof-WTB \citep{dione-2019-developing} treebanks. 
The English GUM treebank \citep{Zeldes2017} was also used to extract examples to illustrate our approach.
Statistics are reported in Table~\ref{table:1}.
 
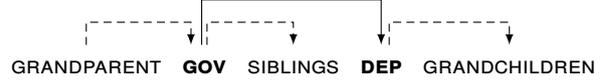
\begin{figure}[t]
\centering
\begin{tikzpicture}[
    every node/.style={
        rectangle,
        inner xsep=0cm,
        inner ysep=0.1cm,
        text height=1.5ex,
        text depth=.25ex,
        node distance=0.3cm
    },
    word/.append style={
        font=\bf\sc\footnotesize,
    },
    deplabel/.append style={
        font=\sc\footnotesize,
        above
    },
    upos/.append style={
        node distance=0cm,
        yshift=0.1cm,
        font=\sc\footnotesize
    }
]
    % He was sitting there, there were two guys sitting
    \node [word] (parent) {grandparent};
    
    \node [word] (gov) [right=of parent,xshift=-0.1em] {\textbf{gov}};

    \node [word] (gov_children) [right=of gov,xshift=-0.1em] {siblings};
        
    \node [word] (dependent) [right=of gov_children,xshift=-0.1em] {\textbf{dep}};
    
    \node [word] (dependent_children) [right=of dependent,xshift=-0.1em] {grandchildren};
    
    \draw [->,densely dashed]
        ($(parent.north)$)
        |-
        ($(gov.north)!0.5!(parent.north)+(0,0.3)$)
        %node[deplabel,xshift=-0.2cm] {deprel}
        -|
        ($(gov.north)+(-5pt,0)$)
    ;
    \draw [->]
        ($(gov.north)+(-1pt,0)$)
        |-
        ($(gov.north)!0.5!(dependent.north)+(0,0.6)$)
        %node[deplabel,xshift=0.2cm] {deprel}
        -|
        (dependent.north)
    ;
    \draw [->,densely dashed]
        ($(gov.north)+(+1pt,0)$)
        |-
        ($(gov.north)!0.5!(gov_children.north)+(0,0.3)$)
        %node[deplabel] {deprel}
        -|
        (gov_children.north)
    ;
    \draw [->,densely dashed]
        ($(dependent.north)+(+3pt,0)$)
        |-
        ($(dependent.north)!0.5!(dependent_children.north)+(0,0.3)$)
        %node[deplabel] {deprel}
        -|
        (dependent_children.north)
    ;

\end{tikzpicture}
\caption{The search space is defined around a governor node (\textsc{gov}) and a dependent node (\textsc{dep}), including the grandparent's governor (\textsc{granparent}) and the governor's other dependents (\textsc{siblings}), as well as the dependent's children (\textsc{grandchildren}).}
    \label{fig:search_space}
\end{figure}

\subsection{Grammatical Rules}\label{sec:grammar_rule}

As stated, a grammar rule is a regular constraint in a language system occurring in a particular context and involving particular subsets of patterns.
Importantly, these rules are probabilistic. In order to build a system capable to extract such rules, we need first to formalize what is a grammatical rule.

In practice, the interpretation of a rule ultimately depends on the theoretical framework adopted.
A framework, in a broad sense, includes the grammatical formalism (\emph{e.g.}\ dependency grammar in our case) %, constituency grammar in \citet{blache-etal-2016-marsagram})
and the annotation scheme (\emph{e.g}.\ UD, SUD, etc.)
How the linguistic hypothesis and question are formulated, and which grammatical features are used to answer to the linguistic question also depend on this.
Our proposed rule extraction formalization is easy to generalise to any theoretical framework, independent of the rule extraction method as well as it is agnostic to the language treebank used.

We first give an intuitive definition of what we call a syntactic grammar rule in this work, based on the example of the subject position in Figure~\ref{fig:subj_parataxis}:
\begin{enumerate}
    \item Firstly, we need to define what we are looking at, \emph{e.g.}\ all dependency relations or a specific subset of them? The scope \textsc{S} of the rule is a given pattern and we consider all dependencies satisfying \textsc{S}.
    %In this paper, \textsc{S} is an unlabeled or a labeled dependency relation, \emph{e.g.}\ the subject relation in the rules for subject inversion in section~\ref{sec:intro}.  
    %We say that the scope of the rule is all relations satisfying a given pattern \textsc{S}.
    In the example, this is all relations of type subject.
    \item Secondly, we need to define what the linguistic phenomenon of interest is. In other words, we seek to identify what triggers satisfaction of response pattern \textsc{Q} amongst all relations that satisfies \textsc{S}.
    %In our examples, we are interested in the order between a dependent and its governor, that is, a subject and the auxiliary or verb governing it.
    In the example, we are interested in subject inversion (subjects that are after their governor in English).
    \item Third and finally, the last element of a rule is the pattern \textsc{P} that acts as a trigger of \textsc{Q} in the scope \textsc{S}. In the example, \textsc{P} would be the fact that the subject has an expletive codependent.
\end{enumerate}

\begin{definition}[Syntactic grammar rule]
Within all dependencies that match a pattern \textsc{S} (the scope),
a grammar rule with frequency $\alpha$ identifies a predictor \textsc{P} that triggers linguistic phenomena \textsc{Q} in $\alpha\%$ of cases.
We denote such a rule by:
{\large\[
%R ~\triangleq~
\textsc{S} \implies (\textsc{P} \overset{\alpha\,\%}{\implies} \textsc{Q})\,.
\]}%
%In particular, when we consider all possible dependencies in a treebank as the scope, \textsc{S} is the empty pattern that matches any input.\sk{ce n'est pas possible de dire ça, ça ne fait pas sens. Notre scope est une dépendance !!!! Même si ici on se limite que à ce cas, on ne peut pas dire que le scope est vide.}
%\cc{Ré-écrire la phrase ci-dessous + expliquer comment lire la formule.}
%In particular, when we consider all possible dependencies in a treebank, we have $\textsc{S} = \top$ the truth predicate.
%pattern-based operational definition of a grammar rule.}
\end{definition}
%\sk{In our case, P1 is at least two nodes linked by a dependency}
%\vspace{10pt}
Our formalization is inspired by correspondence rules in Meaning-text Theory \citep{melcuk1988}.\footnote{In the notation of \citet{melcuk1988}, the rule would be written ``S$\implies$Q\ |\,P'' and read ``S can correspond to Q in the context P'' (the subject is positioned after the verb in presence of an expletive codependent before the verb).}
%\sk{j'ai pas mal changé la rédaction de cette section. On disait 3 fois que S était un pattern et deux autre fois que c'est l'ensemble des matches de S ! Et aussi on limitait ça au cas où S est une dépendance, comme si c'était obligatoire; c'est juste notre choix.}
\section{Rule Extraction Method}

In order to extract rules governing grammatical structures, we first need to define what are the pieces of information that can fill the patterns in these rules.
%In practice, patterns \textsc{S} and its extension \cc{???????!!!!!} \textsc{Q} are manually defined,
In practice, patterns \textsc{S} and \textsc{Q} are manually defined,
as they define the linguistic phenomena of interest in a given scope.
They can therefore rely on any information available in treebanks, without any kind of restriction.

However, potential patterns \textsc{P} are the ones that the machine learning (ML) model must fill.
Therefore, we need to define which are the potential pieces than compose them.
We then explain how we use sparse logistic regression to select and rank them.

\subsection{Features}

Treebanks contain attributes that are used to build features, the input of the ML model.
Remember that what we aim to predict is also an attribute related to a dependency relation satisfying S (Is there a number agreement between the governor and its dependent? Is the dependent after the governor?).

The bigger the search space, the more expressive the pattern \textsc{P} can be.
%\citet{chaudhary-etal-2020-automatic,chaudhary2022autolex} only consider attributes of the governor, the dependent and the relation.
We include high-order features, that is, our search space includes all linguistic information related to the grandparent (governor of the governor), the codependents of the governor and the grandchildren (dependent's children).
Figure~\ref{fig:search_space} illustrates this search space. 

%We use almost all linguistic information encoded in the search space, including relative order between the \textsc{grandparent}, the \textsc{gov} and the \textsc{dep}, and all other lexical, morphosyntactic and syntactic information encoded in the search space. 

Our features are based on all attributes appearing in the search space, with some additions:
\begin{itemize}
    \item We add relative position attributes for the governor and the grandparent, \emph{i.e.}\ one attribute indicating if the governor is before or after the dependent, and one attribute indicating if the grandparent (when existing) is before or after the governor;
    
    \item The \texttt{form} attribute is never used, and we use the \texttt{lemma} attribute only for words from a predefined set of POS tags --- the main motivation is that the large number of values for these attributes introduces noise into the decision process, but that lemma information can be important for closed classes (\emph{e.g.}\ \texttt{AUX} or \texttt{ADP});%\footnote{\label{note:de}For example, in French, adjectives are more likely to precede the noun if the noun has a complement introduced by the adposition \textit{de} ('of', 'from').}
    %This rule is ranked among the first in our experiments on \texttt{ADJ-NOUN} order.}
    \item Although the UD part of speech (\texttt{upos}) tagset is considered  ``coarse'', there are distinctions that may not be meaningful for our purpose --- therefore we add additional attributes indicating if the \texttt{upos} of a word belongs to one of several predefined lists of tags (\emph{e.g.}\ an attribute indicating that \texttt{upos} of the governor is either \texttt{DET} or \texttt{NUM}).
\end{itemize}
We also apply filters on these attributes, especially to prevent information leaks. 
For example, when we are interested in rules governing number agreement, we remove the \texttt{Number} attribute.

A feature is an indicator of the value of one or several specific attributes.
For example, the feature \texttt{DEP.upos=VERB} is equal to 1 if the \texttt{upos} of the dependent (\texttt{DEP}) is \texttt{VERB}, and 0 otherwise.
Features can focus on one attribute (\emph{e.g.}\ \texttt{DEP.upos}) or two attributes (\emph{e.g.}\ \texttt{DEP.upos} and the relative position of its governor).
To limit noise (annotation errors or attributes that are poorly represented in the corpus), we only add features that appear at least 5 times in the scope of the rule.
%\sk{j'ai remplacé unary et binary par simple et double (on peut aussi mettre complexe). Parce qu'un trait binaire, ça évoque autre chose. En plus, on n'utilise pas trop ces termes ensuite.}

%Finally, note that features are hierarchically organized: a feature X can cover a feature Y, meaning that all patterns matching feature Y also match feature X.\sk{ça peut être supprimé, non ?}
%This is illustrated in Figure~\ref{fig:hierarchy}.

%\sk{la section "Agreement and word order rules" qui suivait et a été supprimée me semble pas inutile. }
\begin{comment}

\subsection{Agreement and word order rules}

Once the features defined, we can now give an example for the two types of grammar rules considered: 

\textbf{Agreement rule} There is agreement when two nodes with any constraint and in a dependency relation share the same linguistic attribute. We search P patterns between this to nodes that favors the presence of the same attribute in both nodes,

\[
\begin{array}{c}
\textsc{gov.upos=verb}\\
\textsc{dep.upos=noun}
\end{array}
\Rightarrow (P \Rightarrow \textsc{gov.number=dep.number})
\]

\textbf{Word order rule} Given two nodes in a dependency relation, we look for P patterns that favors the position of one node with respect to the other, 

\[
\begin{array}{c}
\textsc{gov.upos=verb}\\
\textsc{dep.upos=noun}
\end{array}
\Rightarrow (P \Rightarrow \textsc{gov.position=before\_dep})
\]

\end{comment}

\subsection{Sparse Logistic Regression}

We will now describe the method used to select the features that we will use to define the \textsc{P} patterns of the rules.
We formalize the extraction process as \emph{selecting features that allow to identify when a linguistic phenomenon \textsc{Q} is triggered}.
For example, for number agreement between a dependent and its governor, the majority case is defined as ``no agreement or agreement by chance'' and we wish to select features that identify dependencies where number agreement is expected (or forced).
To that end, we rely on a sparse logistic regression.\footnote{We refer the reader to \cite[Sec.~9 \& 25]{shalev2014understandingml} for a more in-depth introduction to sparse logistic regression.}

We denote $F$ the set of features.
Given an input $\vx \in \{0, 1\}^F$ represented by the boolean feature indicator vector,
the decision of a binary linear classifier is defined as follows:
$$
f_{\va, b}(\vx) = \begin{cases}
    1 \quad&\text{if~} \va^\top \vx + b > 0\,, \\
    0 \quad&\text{otherwise,}
\end{cases}
$$
where $\va \in \R^F$ (feature weights) and $b \in \R$ (intercept term) are the parameters.
However, this decision is ``hard'' and therefore inappropriate for our purpose.
Continuing with our example, number agreement can be observed by chance: what we want to learn is the probability to observe an agreement.
To this end, we can instead interpret the predictor as giving the probability that there is a number agreement between the modifier and its governor for input $\vx$:
\begin{equation}
    \textrm{P}(\text{``number agreement''} | \vx)
    =
    \sigma\left(\va^\top \vx + b\right)\,,
    \label{eq:prob}
\end{equation}
where $\sigma(w) = \frac{\exp(w)}{1 + \exp(w)}$ is the sigmoid function that maps a real value to a value in $[0, 1]$.
As such, we seek to identify features that drive this probability closer to one (forced agreement) compared to the majority case (where an agreement may happened by chance in roughly $50\%$ of observations).

By definition of dot product $\va^\top \vx = \sum_{f \in F} \eva_f \evx_f$, a feature $f \in F$ contributes to the probability in Equation~\ref{eq:prob} if and only if its associated weight $\eva_f$ is not equal to 0. Intuitively, as the sigmoid function is strictly increasing, a positive $\eva_f$ (resp.\ negative $\eva_f$) means that the presence of this feature will increase the probability of observing a number agreement (resp.\ a disagreement).\footnote{Note that there are limitations to this interpretation that will be discussed in Section~\ref{sec:rule_analysis}.}

To learn parameters $\va$ and $b$ of the model, we rely on the following standard training objective~:
\begin{equation}
\min_{\va \in \R^F}\quad
%\underbrace{\frac{1}{|D|}\sum_{(\vx, y) \in D} \ell\left(\va^\top \vx + b, y \right)}_{= L(\va, b; D)}
\frac{1}{|D|}\sum_{(\vx, y) \in D} \ell\left(\va^\top \vx + b, y \right)
~~+~~
\lambda r(\va)\,,
\label{eq:opt}
\end{equation}
where $D$ is the training dataset (the scope of the rule), $\ell$ is a loss function, $r$ is a regularization term and $\lambda \geq 0$ is the regularization strength.
In particular, we choose:
\begin{itemize}
    \item $\ell(w, y) = - yw + \log(1+\exp w)$, the negative log-likelihood, which is well-calibrated for probability estimation \cite{reid2010binarylosses};
    \item $r(\va) = \| \va \|_1 = \sum_{f \in F} |\eva_f|$, the L1-norm which induces parameter sparsity, \emph{i.e.}\ depending on the value of $\lambda$, more or less values in $\va$ will be exactly equal to 0 \cite{bach2012optimization}.
\end{itemize}
Importantly, the intercept term $b$ is not regularized.
Applying regularization to $b$ is equivalent to assuming a Laplacean prior with mean $0$ \cite[Sec.~13.3]{murphy2012ml}, which may be unsuitable.\footnote{For example, in the case of number agreement, regularizing $b$ is equivalent to postulating that there is a $50\%$ chance of observing an agreement. Without regularization, we leave full flexibility to the model to estimate this prior to agreement.}
For unregularized $b$, when $\lambda \to \infty$, all feature weights will be null, \emph{i.e.}\ $\va = 0$, and the intercept $b$ will capture the mean probability of phenomena \textsc{Q} in the scope:
$$
\sigma(b)
=
\frac{1}{|D|}\sum_{(\vx, y) \in D} y\,.
$$

\textbf{Practical consideration.}
The optimization problem in Equation~\ref{eq:opt} is challenging as there is a large number of parameters and the regularization term is non-smooth.
The \textsc{ScikitLearn} library relies on \textsc{Liblinear} for this problem, but does not support unreguralized intercept term. %\footnote{\url{https://scikit-learn.org/stable/modules/generated/sklearn.linear_model.LogisticRegression.html}}
In this work, we therefore use \textsc{Skglm} \citep{skglm}.
%, which was designed for these types of problems and support unreguralized intercept.

\subsection{Ranking via Regularization Path}
\label{sec:reg_path}

The term regularization path refers to the function $\lambda \mapsto \widehat\va(\lambda)$ that maps the regularization parameter $\lambda$ to the model parameters $\va$ minimizing the training problem (Eq.~\ref{eq:opt}), denoted $\widehat\va(\lambda)$.
We say that feature $f \in F$ is in (resp.\ out) the regularization path at $\lambda$ if $\widehat\eva(\lambda)_f \neq 0$ (resp.\ $=0$).
Importantly, when $\lambda \to \infty$, $\widehat\va(\lambda)_f = \vzero$, \emph{i.e.}\ no feature is selected by the classifier.

To select and rank features that form pattern \textsc{P}, we proceed as follows:
\begin{enumerate}
    \item We fix $\lambda^{(0)}$ to a reasonable value where $\widehat\va(\lambda^{(0)}) = \vzero$;
    \item As we cannot explore all possible $\lambda$ values, we fix a number $k$ of additional values we will inspect;
    \item We fix $\lambda^{(k)}$ to a reasonable minimum value for the regularization strength.
\end{enumerate}
Then we train the model for $k+1$ regularization strengths $\lambda^{(i)}$, $0 \leq i \leq k$, evenly distributed between $\lambda^{(0)}$ and $\lambda^{(k)}$.
In practice, we fix $k=100$, $\lambda^{(0)} = 0.1$ and $\lambda^{(k)} = 0.001$.

We say the feature $f \in F$ enters the regularization path at step $i$ if and only if $\forall j \in \{0,...,i-1\}: \widehat{\eva}(\lambda^{(j)}) = 0$ and $\widehat{\eva}(\lambda^{(i)}) \neq 0$.
In other words, $f \in F$ enters the regularization path at step $i$ if it is the first step where its associated weight in the parameter vector $\va$ is non-null. %\footnote{A feature can also ``leave'' (temporarily) the regularization path. However, this rarely happens in practice and it can be ignored.}
The features are ranked by entering order in the regularization path:
if features $f$ and $f'$ enter the path at steps $i$ and $j$, respectively, such that $i < j$, we consider that feature $f$ is more salient than feature $f'$ as $f$ is more important to predict the frequency of pattern \textsc{Q} in the scope defined by \textsc{S}.

\textbf{On hyperparameters.}
Importantly, note that the set of considered regularization strengths is the only hyperparameter of our model.\footnote{More precisely, the model also have hyperparameters related to the optimization algorithm. However, they can be safely be ignored and set to default values.}
This is a very important feature as it is difficult to tune hyperparameters.
Indeed, the standard ML methodology consists of tuning them on a development set.
However, predictions of the classifier are frequencies: they should be interpreted, given a set of observed features, as the frequency of the targeted phenomenon.
If the model predicts 45\% of agreements, this mostly means agreements by chance, and it does not make sense to consider that the model makes an error if for a given data point $\vx$, the output label is $1$.
As product of features can be rare, especially for not-so-frequent linguistic phenomena, the accuracy variance will be high on a small development set, making the hyperparameter tuning procedure meaningless.
Our approach does not suffer from such limitation.

\subsection{Rule Analysis}\label{sec:rule_analysis}

It is important to note that, contrary to popular belief, the parameters of a linear model are not directly interpretable as explanatory factors.
As explained in previous section, the regularization path only gives us ranking of features (by their saliency), but the particular non-null values in $\va$ must be ignored due to potential non-linear correlations \cite{achen2005garbagecan}.
Therefore, for each considered feature for pattern \textsc{P}, we must:
(1) check if it is a trigger of \textsc{Q} or $\neg\textsc{Q}$ and (2)~compute other statistical measures to provide more insight.

In the following, we denote $\#(P)$ the number of dependencies matching a given pattern $P$.

\textbf{Statistical test.}
We denote $\mu$ the Bernoulli parameter of the expected distribution,
that is, the ratio of pattern \textsc{Q} in the scope:
\[
\mu = \frac{\#(\textsc{S}~\land~\textsc{Q})}{\#(\textsc{S})}\,.
\]
Consider a potential rule $\textsc{S}\Rightarrow(\textsc{P}\overset{\alpha\,\%}{\Rightarrow} \textsc{Q})$ where:
\[
\alpha = \frac{\#(\textsc{S}~\land~\textsc{P}~\land~\textsc{Q})}{\#(\textsc{S}~\land~\textsc{P})}\,.
\]
This grammatical rule is of statistical interest if $\alpha$ is ``different enough'' from $\mu$.
However, given that we approximate $\alpha$ using a limited number of samples $n$,
we need to test if this divergence is statistically significant.
The G-test statistic is defined as follows:
\begin{equation}
G = 2 \times n \times \left( \alpha \cdot \ln\frac{\alpha}{\mu} + (1-\alpha) \cdot \ln\frac{1 - \alpha}{1 - \mu} \right)\,.
\label{eq:gtest}
\end{equation} %Similar to ${\chi}^2$ hypothesis test, it is more robust for small samples.
Note that this is simply a KL-divergence between Bernoullis parameterized by $\mu$ and $\alpha$ but reweighted by the number of samples used to estimate $\alpha$.
The p-value of a G-test is calculated by looking at the tail probability of the $\chi^2$ distribution with the right degrees of freedom (\emph{i.e.}\ 1 in our case). 
%This test has a better an approximation to the $\chi^2$ distribution than the $\chi^2$ test \citep{harremoes-chi2-2012}.
%The direction of the statistical significance (positive or negative significance) is interpreted by the sign of the difference between observed and expected occurrences.
We consider a grammar rule to be statistically significant if $p < 0.01$, \emph{e.g.} agreement is not by chance if it is statistically significant (see \citet{chaudhary-etal-2020-automatic}). % under the independence hypothesis, filtering out less significant ones.%\footnote{For small samples, Fisher's exact test is preferable. The experiments in this paper yield exactly the same results for both tests. We choose to use the G-test because its statistic is more fine-grained than using the $log10$ of the p-value to indicate significance.}
Note that the G-test statistic can be adapted for rules triggering $\neg \textsc{Q}$.
Moreover, it can also be used to rank rules:
as it is (mostly) a KL-divergence, rules that deviate more from the base distribution $\mu$ will have a higher value.
%We use the G-statistic, as only a one way to rank the potential grammar rules. A pattern with a higher G-statistic is more likely to be a grammar rule.

\textbf{Association measures.}
% We calculate the effect size \textit{Odds Ratio (OR)}. It quantifies the odds of the linguistic phenomenon $P2$ of interest in a set of patterns to the odds ($Q$) of it occurring in another set ($\neg Q$). ORs less than 1 indicate a negative association between patterns. We use it to compare patterns across treebanks, as it is a standardized measure. In addition, 
We compute the \textit{Cramer's phi $\phi_c$} effect size using the G-test statistic. Unlike \citet{chaudhary-etal-2020-automatic}, we do not use it to filter rules because the threshold considered for doing so vary depending on the data and domains. However, we use effect size as another interpretable measure to rank rules.

\textbf{Coverage and precision.}
We hypothesize that reliable grammar rules are those that covers largely the linguistic phenomenon of interest and those that are precise. Coverage/recall and precision of the pattern $P$ are defined as follows:

\vspace{10pt}

\begin{tabular}{@{}lcc@{}}
   \textbf{Rule}{\phantom{\huge\textbf{P}}} & \textbf{Coverage} & \textbf{Precision} \\
   \midrule
   {\small$\textsc{S} \implies (\textsc{P} \overset{\alpha\,\%}{\implies} \textsc{Q}$)}
   &
   $\frac{\#(\textsc{S} \land \textsc{P} \land \textsc{Q})}{\#(\textsc{S} \land \textsc{Q})}$
   &
   $\frac{\#(\textsc{S} \land \textsc{P} \land \textsc{Q})}{\#(\textsc{S} \land \textsc{P})}$
   \\
   {\small$\textsc{S} \implies (\textsc{P} \overset{\alpha\,\%}{\implies} \neg\textsc{Q}$)}
   &
   $\frac{\#(\textsc{S} \land \textsc{P} \land \neg\textsc{Q})}{\#(\textsc{S} \land \neg\textsc{Q})}$
   &
   $\frac{\#(\textsc{S} \land \textsc{P} \land \neg\textsc{Q})}{\#(\textsc{S} \land \textsc{P})}$
\end{tabular}
\vspace{10pt}

Note that these measures must be used with cautions.
For example, a number agreement between a dependent and its governor can happen by chance.
As such, a rule that trigger forced number agreement will never achieve a $100\,\%$ cover (and will probably be way lower).

\section{Experimental Results}\label{sec:results}

\begin{table*}[t]
\centering
\small
\begin{tabular}{lllcrr}
\toprule
& \textbf{\textsc{P pattern}} & \textbf{position} & \bm{$\lambda$} & \textbf{coverage} & \textbf{precision}\\
\midrule
1 & dep.NumType=Ord,dep.rel\_synt=mod & before & 0.042 & 21.3 & 98.2\\   
2 & dep.rel\_synt=mod,siblings.upos=ADP & before & 0.033 & 40.9 & 39.9\\
3 & grandchildren.upos=ADP & after & 0.022 & 12.8 & 98.9\\
4 & gov.upos=NOUN|PROPN|PRON,grandchildren.upos=ADP & after & 0.021 & 12.8 & 98.9\\
5 & dep.rel\_synt=mod,siblings.lemmas=de & before & 0.012 & 31.8 & 41.6\\
6 & grandparent.position=before\_gov,grandparent.upos=ADP & after & 0.011 & 49.9 & 75.1\\
7 & dep.Degree=Cmp,dep.rel\_synt=mod & before & 0.011 & 5.9 & 90\\
8 & grandparent.lemma=de & after & 0.01 & 23.8 & 79.7\\
9 & gov.upos=NOUN|PROPN|PRON,grandparent.lemma=de & after & 0.01 & 23.8 & 79.7\\
10 & gov.rel\_synt=comp:obj,siblings.Definite=Def & after & 0.01 & 35.1 & 74\\
\bottomrule
\end{tabular}
\caption{The ten most salient noun-adjective word order rules, extracted from the SUD\_Spanish-AnCora treebank and ranked based on the order given by the linear classifier. \textsc{P} is the pattern that favors the adjective position indicated in the second column. See Section~\ref{sec:reg_path} for the interpretation of $\lambda$. Coverage and precision are expressed as percentages.} \label{table:2}
\end{table*}

The extracted rules for word order are based on the relative position of a governor and its dependent. 
We look at particular cases, such as the \textsc{subj} relation and the relative position of the subject regarding its verb/auxiliary governor. For example, we obtained patterns that favor placing the subject before or after the verb, compared to the general positioning of the subject. 
%We look at the general case with an unlabeled dependency and the position of any dependent regarding its governor, as well as particular cases, such as the \textsc{subj} relation and the relative position of the subject regarding its verb/auxiliary governor. In the general case, we get the most salient word order rules, i.e. patterns that favor a particular word order over the general. 
%The second case gives us patterns that favor placing the subject before or after the verb, compared to the general positioning of the subject. The two results are different for a language such as English, which is weakly head-initial \citep{Gerdes_Kahane_Chen_2021}, while the subject relation is strongly head-final.
%The two results are different for a language such as English, which is weakly head-initial, while the subject relation is strongly head-final.\citep{Gerdes_Kahane_Chen_2021}.
%The extracted rules for word order are based on the relative position of any pair of connected nodes, as well as on the position of the governor and dependent in a \texttt{subj} and \texttt{comp:obj} dependency relations.\footnote{We included the verb as the governor for \texttt{comp:obj}  because this relation is also used to connect adpositions with their dependents in the SUD framework.}
We partly follow \citet{chaudhary2022autolex}, which in turn based their work on WALS word order rules  \citep{wals}. We extract word order rules for the non-dominant position for pairs subject-verb/auxiliary, object-verb and adjective-noun. As for the agreement rules, we explore agreement in number, person and gender between a governor and a dependent having these features, as well as agreement for particular relations, such as \textsc{subj} or \textsc{comp:obj}.
%This can be done for any treebank and has been applied to the 259 SUD treebanks of the 148 languages of UDv2.13.
%The results are available on {\url{https://autogramm.github.io/grex-LREC2024}.} \sh{for at least one treebank of each language? C'est plus realiste}\sk{autant le faire pour les 259 treebanks, non ? c'est ce que font Chaudary et al.} \sh{ils font seulement 78 langues }
The code to reproduce the experiments is freely available,\footnote{\url{https://github.com/FilippoC/grex-lrec-coling-2024}} together with a tool accessible online to visualize results on different treebanks.\footnote{\url{https://autogramm.github.io/grex-lrec-coling-2024}}

In the following, we analyze in detail, as an exploratory example, the ten most salient word-order rules between a noun and its dependent adjective in Spanish (Table~\ref{table:2}). We will then provide a study on word order involving the object in Wolof (Table~\ref{table:3}). Finally, we comment on the extracted agreement rules. As far as we know, the rules analysed have not been found in comparable previous works, unless otherwise stated. The following qualitative analysis covers only a small portion of the rules that were extracted, but it is intended to be representative of the rules extracted by our method. No filters or pruning techniques were applied. Interesting and precise grammar rules and patterns are found for all the languages studied.

\subsection{Noun-adjective order in Spanish}

In Spanish, the dominant order is for the adjective to follow its governing noun. In the SUD\_Spanish-AnCora treebank, about 28\% of dependent adjectives precede their noun. We search for rules that, for all noun-adjective in a dependency relation with a governing noun (\textsc{S} \textsc{gov.upos=noun,dep.upos=adj}), trigger the adjective anteposition (\textsc{gov.position=after\_dep}) and more exactly a shift in distribution.

%Adjective position is primarily determined lexically, and \citet{chaudhary2022autolex} had captured this factor.\footnote{Their rules are based on lexical attributes, such as being an ordinal or having such and such lemmas.} But there may be specific morphosyntactic properties and features influencing the preceding position. Therefore, we search for rules that, for all noun-adjective in a dependency relation with a governing noun (\textsc{S} \textsc{gov.upos=noun,dep.upos=adj}), trigger the adjective anteposition (\textsc{gov.position=after\_dep}).\footnote{\sh{P patterns in Table \ref{table:2} are described as follows: each node in our search space has a linguistic attribute which name is introduced by a dot (\textsc{.NumType}, \textsc{.rel\_synt}), followed by its corresponding value (\textsc{=Ord}, \textsc{=mod}). A feature can have up to two linguistic attributes.}}

 Adjective position is primarily determined lexically and our most salient rule (rule 1 in Table~\ref{table:2}) indicates that ordinal adjectives \ref{ex:ordinal-example} in a modifier dependency relation precede almost always (precison 98\%) their noun, while adjectives in general precede their noun only 28\% of cases. It's worth noting that rule 1 covers about 21\% of the adjectives that precede the noun in the treebank. This is not the only rule that includes a lexical feature: rule 7 shows that comparative adjectives (\textsc{dep.degree=cmp}) are also much more likely to be before their noun.

\exg.
    Desde un \textbf{sexto} piso\label{ex:ordinal-example}\\
    from a \textbf{sixth.{\sc \textbf{ord}}} floor.{\sc noun}\\
    `From a sixth floor' 

However, contrary to previous works (notably \citet{chaudhary2022autolex}), we also capture pure syntactic rules (\emph{i.e.}\ not involving lexical attributes). In the ten most salient rules, the majority of them involve the presence of prepositional phrases (PP) in some way (\emph{cf.}\ ADP for adposition). Rule 2, for example, indicate that nouns having a PP tend to have more preceding adjectives \ref{ex:prep-phrase}. Rule 5 is a finer-grained version of this rule indicating that nouns governing a preposition \textit{de} ('of', 'from') have also more preceding adjectives than the expected. The syntactic interpretation of rules such as these must be done with caution. The anteposition may be explained by the fact that the postnominal position is already occupied by the PP, but also by the fact that the noun with a PP prefers adjectives that are usually placed before it, like ordinals. These two rules can be considered as syntactic tendencies due to their low/medium precision. They appear in the top rules due to their significant coverage.

\exg.
    \textbf{Rara} y triste evidencia \textbf{de madurez}\label{ex:prep-phrase}\\
    \textbf{Rare} and sad evidence \textbf{of.{\sc \textbf{adp}} maturity}\\
    `Rare and sad evidence of maturity.' 

Rule 6 (and its finer-grained versions, rules 8 and 9) shows that adjectives follow their noun in almost 75\% of cases when the latter is governed by a preposition (\textsc{grandparent.upos=adp}) that is, \emph{i.e.} when they are in a PP. Again, nouns may prefer adjectives in a post-nominal position than preceding them. 

Finally, we capture also really precise rules for the post-position of the adjective. Dependent adjectives governing a preposition (\textsc{grandchildren.upos=adp}), that is, which have a PP, are more likely to be postponed (rules 3, 4). We find the same rule in English, where the dominant order is adjective-noun. In these cases, the postposition of an adjective is explained by the weight or heaviness of the syntactic group they govern, see \citet{thuilier2012contraintes} for French. 

Our method can capture these rules due to the a large search space and sensitivity to distribution shifts. In addition, even though we proposed a more expressive system than previous works, our rules remain easy to interpret. However, the interpretation of the rules must be lexico-syntactic and must take into account the probabilistic and overlapping nature of the extracted rules. 

\subsection{Object order in Wolof}\label{sec:wolof_results}

\begin{table*}[t]
\centering
\small
\begin{tabular}{lllcrr}
\toprule
& \textbf{\textsc{P pattern}} & \textbf{position} & \bm{$\lambda$} & \textbf{coverage} & \textbf{precision}\\
\midrule
1 & dep.upos=PRON,gov.VerbForm=Fin & before & 0.1 & 92.6 & 63.2\\   
2 & dep.upos=NOUN|PROPN|PRON,dep.upos=PRON & before & 0.077 & 94.7 & 58.9\\
3 & dep.upos=PRON & before & 0.067 & 94.7 & 58.9\\
4 & dep.upos=PRON,gov.Mood=Ind & before & 0.064 & 91.7 & 63.4\\
5 & dep.upos=PRON,gov.rel\_deep=mod@relcl & before & 0.041 & 36.7 & 96.7\\
6 & dep.upos=PRON,gov.rel\_synt=comp:aux & before & 0.033 & 47 & 72.1\\
11 & dep.upos=PRON,grandparent.lemma=di & before & 0.025 & 30.3 & 85\\ 
12 & dep.Case=Acc,gov.rel\_synt=comp:aux & before & 0.023 & 35.1 & 78.8\\
27 & dep.Case=Acc,gov.Aspect=Perf & before & 0.009 & 8.2 & 92.5\\
54 & dep.upos=PRON,gov.Aspect=Perf & before & 0.003 & 9.1 & 85.7\\
\bottomrule
\end{tabular}
\caption{Word order rules for the pre-verbal position of the object mentioned in \ref{sec:wolof_results} and extracted from the SUD\_Wolof-WTB treebank. The order is given by the linear classifier, the last rule of the table coming at position 54. See Section~\ref{sec:reg_path} for the interpretation of $\lambda$. \textsc{P} is the pattern that favors the object anteposition. Coverage and precision are expressed as percentages.} \label{table:3}
\end{table*}

\citet{chaudhary2022autolex} extract three word order rules for the object in Wolof, which are our rules 5, 12, and 54 of Table~\ref{table:3}: the object tends to be placed before the verb when it is an accusative pronoun governed by a verb depending on an auxiliary (rule 12), or a pronoun governed by a verb being in a relative clause (rule 5), or bearing the perfective aspect marker (rule 54).\footnote{Our rule 12 does not indicate that the dependent is a pronoun, but only pronouns have a Case feature in the Wolof-WTB treebank.} They do not capture more general tendencies, neither more specific rules. 

Our set of rules can be interpreted as a hierarchy of rules, where the three previous rules are particular nodes. Rule 2,3 indicate that pronouns tends to be placed before the verb (contrary to nouns). Rules 1 and 4 shows that this tendency increases if the governor is a finite verb or if it is a verb in the indicative mood. Rule 6 indicates that the tendency also increases if the verb depends on an auxiliary (precision 72\%). It increases even more if the auxiliary is \textit{di} (rule 11, precision 85\%) or if the pronoun is marked by the accusative case (rule 12, precision 78\%). Even if rule 12 is descriptive by itself, we find more interesting to have it in relation with rules 6 and 11, which are more covering for the first and more precise for the second. Similarly, rule 54 for verbs with a perfective aspect is interesting, but rule 27 adding the constraint for the pronoun to be accusative is more precise (92\% vs 85\%). It is not clear why \citeauthor{chaudhary2022autolex}'s tool add this additional condition for object pronouns depending on a verb and on an auxiliary (and choose rule 12 rather than rule 6), and not for pronouns depending on a verb with a perfective aspect (rule 54 rather than rule 27).

Having this rich hierarchy of rules (and tendencies) is invaluable, but it also raises the question of the selection of particular rules. We are not sure that a unique and simple solution can be provided and the answer probably depends on the objective. We can be more interested by results with a high precision (they are the cases where we can really talk about “rules”) or prefer a compromise between precision and coverage, which is what our method seems to do (see Section~\ref{sec:ranks}). If we do not keep the whole hierarchy of results, it is also important to select a set of rules/tendencies that covers most of the data and are not too redundant, which is what \citeauthor{chaudhary-etal-2020-automatic}'s method based on decision trees does well.

\subsection{Agreement}

%We search agreement rules in number, person and gender between two nodes having these features. We tested several ways of capturing agreement rules using variants in the \textsc{S} scope. We extract rules specifying the \textsc{upos} of the governor and the dependent (e.g. agreement between a noun and an adjective). We also look for agreement with subjects and objects. Finally, we also extract rules without any other constraint in the \textsc{S} scope. 

Agreement is a phenomenon that is more restrictive than word order, really stable in Indo-European language such as English, French and Spanish. This is the reason why the rules extracted are less diverse and more expected. For number agreement, we capture usual rules like agreement between a noun and a determiner (\emph{e.g.} rule 17 in Spanish) or the assessments that dependent adjectives agree with their governor and that verbs agree with their subjects (\emph{e.g.} rule 4 and 41 in French). As regards negative rules, we exclude, for example, agreement between objects and their verbs in French, a crucial test for our method, which represent ~3\% of mismatch between heads and dependents having the number feature.   

The extraction of gender agreement rules suffers from different annotation strategies. Being an inherent property of the noun or the referred entity, it may not always be expressed by morphosyntactic flexion. In any case, principal agreement rules were found in French and Spanish, as the agreement between the noun and its adjectives and determiners.

The most informative way to extract agreement patterns is by checking for agreement between any head and dependent, as its done in \citet{chaudhary-etal-2020-automatic}. This provides a general overview of the agreement system for a given language. 
%The majority of agreement rules we already mentioned are also found using this initial pattern. 
However, due to its under specification, some rules are more general. For example, we extract rules expressing the fact that number and gender agreement co-occur both in French and Spanish (see French rules 28 and 29 for number agreement). Our findings also indicate that in French and Spanish, number agreement generally occurs when the governor follows the dependent. This is particularly evident in the agreement between determiners and nouns, which is the largest set of agreed pairs in both languages. On the contrary, about 85\% of dependents located after their head, if they have a number feature, do not agree in French (see rule 19). Global analysis like the precedent one are useful in quantitative typology \citep{Gerdes_Kahane_Chen_2021}.

\subsection{Ranks}\label{sec:ranks}

By expanding the search space and due to the nature of our rules, we extract a greater number of rules compared to previous works. It is why order and ranking the rules is crucial to achieving better interpretability. Beyond the order given by the linear classifier (section~\ref{sec:reg_path}), it is possible to rank the rules using the other statistical measures described (section~\ref{sec:rule_analysis}), as is done in Corpus Linguistics (e.g. \citealp{pecina-schlesinger-2006-combining}), prioritizing differences of aspects of the same rule, like the precision or coverage of the rule. In addition, these measures could be used to prune subsets in different ways and reducing the final number of rules.

We compared the orders of the rules given by the model and by using the G-test statistic. We computed a Spearman's rank order correlation for word order experiments, and we obtained significant results, going from weak-moderate to high positive correlation ($r > 0.40, p~value < 0.01$) with some exceptions (\emph{e.g} adjective-noun word order). This could mean that while they are in most cases order is correlated, they do not highlight exactly the same type of information. It is worth noting that the order given by the classifier is partial and less granular, with several rules occupying the same place.

\section{Conclusion}

We proposed a novel method for grammatical rule extraction from treebanks.
Our approach is based on (1) a formal definition of what is a syntactic grammatical rule and (2) the regularization path of a sparse logistic classification model.
Moreover, we showed in our result analysis that it is important to capture less pronounced shift distributions to include a greater variety of rule patterns.
This allows to increase the expressiveness of rules and to capture linguistic phenomena that are out of the scope of previous work \cite{chaudhary-etal-2020-automatic,chaudhary2022autolex,blache-etal-2016-marsagram}.

Alongside our main contribution, we hope that our clear formal description of each component will help to develop future work on the topic.

\begin{comment}
Extraire des règles à partir d'un grand nombre de traits pose des problèmes d'explosion combinatoire si on commence à regarder toutes les combinaisons de traits. Ceci nous a amené à nous tourner vers des méthodes qui n'avaient pas encore été explorées pour l'extraction de règles de grammairre à partir de treebanks, à savoir sparse logistic regression <@caio description de la méthode en qq mots>. En plus de donner des résultats satisfaisants, la méthode s'avère particulièrement efficace. Il y a bien sûr des limites à l'extraction de grammaires à partir de treebanks, essentiellement à cause de certains choix d'annotation et au manque d'informations cruciales.\footnote{Par exemple, on sait que l'ordre des mots est fortement conditionnée par l'information packaging, information dont nous ne disposons pas.} Une de nos piste de recherche future est d'enrichir automatiquement le treebank avec d'autres couches d'information, comme des classes sémantiques.
\sk{je m'arrêterai là-dessus et laisserai tomber la suite
}
\end{comment}

\section{Limitations}

The extracted grammatical rules rely on the annotation scheme. Therefore, some patterns extracted may reflect corpus properties and theoretical decisions more than actual grammar rules. Our search for agreement rules, which is consistent with previous researches, is limited to syntactic agreement. As a result, one of the most salient (negative) rule for the subject agreement is to exclude subjects having coordination, which is due to the fact that the verb is linked to the first conjunct, not reflecting the features of the coordinated phrase. While our extraction system does not capture this kind of semantic agreement, its behavior remains consistent and contributes to understandability.

Grammar rules also reflect annotation errors or lack/excess of annotation. The most precise negative rule for number agreement in French captures full dates (\textit{6 janvier 1902} ‘January 6, 1902') because the year is labeled as plural while its head, the month, as a singular noun.\footnote{Note that \textit{1902} when used as a proper name as here should be singular, since it triggers a singular agreement: \textit{1902 is …}} 
%Even if the linguistic value of a pattern is not explicit, a significant pattern highlights consistent constructions. E.g. relatives in Spanish are annotated using an underspecified dependency relation (\texttt{acl} in UD, instead of \texttt{acl:relcl}). 
%This is why the interpretation of the rule that the subject's post-verbal position is more probable due to the presence of a relative is not easy. \sh{je ne sais pas si c'est interessant}\sk{je ne pense pas que ce soit un problème puisque quasiment tous les verbes acl qui ont un sujet sont têtes d'une relative à part "le fait que Zoé vienne" ou "l'envie que Zoé a de venir". Donc supprime.} Associate queries to treebanks is a way to know what each rule is matching.\footnote{In this sense, a secondary utility of our system is error mining, as systematic errors resulting from parsers or manual annotation are statistically salient.}

% (\texttt{head\_rel\_shallow=mod,gp\_upos=NUM}) 

Extracted rules for marginally existing linguistic phenomena are much more loose and less significant. This is the case of agreement between objects and verbs in the languages studied. The observed results show preferences for language use, not grammar rules, such as the fact that singular objects are significantly paired with singular subjects (likewise for plurals), resulting in a tendency for the verb to agree with its object. Conversely, the agreement rule of a past participle with its object when the object precedes it existing in French was not found, probably because the relative pronoun \textit{que} (‘that') does not bear any agreement feature.

\section*{Acknowledgement}
We thank Mathurin Massias and Badr Moufad for the help with the \textsc{Skglm} library.
We also thank Bruno Guillaume, Guillaume Bonfante, and Guy Perrier for their insightful conversations and feedback.
We thank the anonymous reviewers for their comments and suggestions.
All authors are supported by a public grant overseen by the French ANR (Autogramm, ANR-21-CE38-0017).

\section{Bibliographical References}\label{sec:reference}
\bibliographystyle{lrec-coling2024-natbib}
\bibliography{lrec-coling2024-example}

%\section{Language Resource References}
%\label{lr:ref}
%\bibliographystylelanguageresource{lrec-coling2024-natbib}
%\bibliographylanguageresource{languageresource}

\end{document}